%% file: main.tex
\documentclass[10pt,twocolumn,letterpaper]{article}

\usepackage[pagenumbers]{arxiv}

\usepackage{graphicx}
\usepackage{amsmath}
\usepackage{booktabs}
\usepackage{times}
\usepackage{epsfig}
\usepackage[font=small,skip=5pt]{caption}
\usepackage{multirow}
\usepackage[hidelinks]{hyperref}

\input{notation}

\input{math_commands}

\usepackage[capitalize]{cleveref}

\begin{document}

\title{Control-NeRF: Editable Feature Volumes for Scene Rendering and Manipulation}

\newcommand*\samethanks[1][\value{footnote}]{\footnotemark[#1]}

\author{Verica Lazova$^1$$^2$ \\
{\tt\footnotesize verica.lazova@uni-tuebingen.de}
\and
\hspace{2em} Vladimir Guzov$^1$$^2$ \\
\hspace{2em} {\tt\footnotesize vladimir.guzov@mnf.uni-tuebingen.de}
\and
Kyle Olszewski$^3$ \\
{\tt\footnotesize kolszewski@snap.com}
\and
\and
\and
Sergey Tulyakov$^3$ \\
{\tt\footnotesize stulyakov@snap.com}
\and
Gerard Pons-Moll$^1$$^2$ \\
{\tt\footnotesize gerard.pons-moll@uni-tuebingen.de}
\and
\vspace{1cm}
\and
\normalsize $^1$University of Tübingen, \hspace{0.6cm} $^2$Max Planck Institute for Informatics, Saarland Informatics Campus, \hspace{0.6cm} $^3$Snap Inc.}

\twocolumn[{
\renewcommand\twocolumn[1][]{#1}

\maketitle
\vspace{-1cm}
\begin{center}
\newcommand{\teaserwidth}{\textwidth}
\centerline{\includegraphics[width=\teaserwidth,clip]{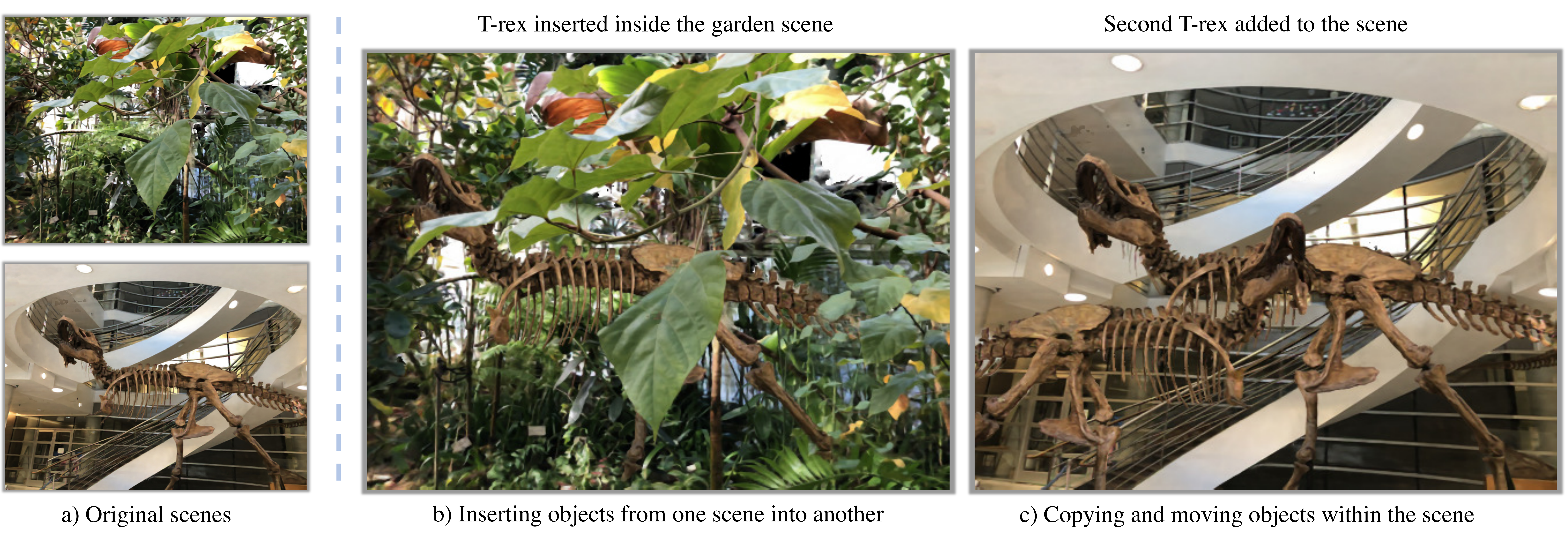}}
\captionof{figure}{Our model allows novel view synthesis and controlled scene editing and manipulation: editing parts of the object applying rigid and non-rigid transformations(a); adding and removing objects in the scene (b); moving objects from one scene into another (c).}
\label{fig:teaser}
\end{center}
}]

\maketitle

\input{sections/abstract}

\input{sections/introduction}

\input{sections/related_work}

\input{sections/method}

\input{sections/experiments}
\input{sections/conclusion}

\clearpage

\bibliographystyle{ieee_fullname}
\bibliography{references}

\clearpage

\setcounter{section}{0}
\renewcommand\thesection{\Alph{section}}

\input{supp_sections/training}
\input{supp_sections/ablations}
\input{supp_sections/comparisons}

\clearpage

\end{document}

%% file: notation.tex
\usepackage{xspace}

\makeatletter
\DeclareRobustCommand\onedot{\futurelet\@let@token\@onedot}
\def\@onedot{\ifx\@let@token.\else.\null\fi\xspace}

\def\eg{\emph{e.g}\onedot} 
\def\ie{\emph{i.e}\onedot}

\makeatother

\newcommand{\modeltheta}{\mathrm{\Theta}}

\newcommand\reallywidehat[1]{%
\savestack{\tmpbox}{\stretchto{%
  \scaleto{%
    \scalerel*[\widthof{\ensuremath{#1}}]{\kern-.6pt\bigwedge\kern-.6pt}%
    {\rule[-\textheight/2]{1ex}{\textheight}}%
  }{\textheight}%
}{0.5ex}}%
\stackon[1pt]{#1}{\tmpbox}%
}

\newcommand{\norm}[1]{\left\lVert#1\right\rVert}

\newcommand{\ray}{\mathbf{r}}
\newcommand{\Ltrain}{\mathcal{L}}
\newcommand{\raybatch}{\mathcal{R}}
\newcommand{\Ccoarse}{\hat{C}_c(\ray)}
\newcommand{\Ccoarsefull}{\hat{C}_c(\ray,V_s,\modeltheta)}
\newcommand{\Cfine}{\hat{C}_f(\ray)}
\newcommand{\Cfinefull}{\hat{C}_f(\ray,V_s,\modeltheta)}
\newcommand{\Ctrue}{C(\ray)}

\newcommand{\trainsceneimages}{\mathcal{I}}
\newcommand{\trainscenevolumes}{\mathcal{V}}
\newcommand{\testsceneimages}{\mathcal{G}}

\newcommand{\timenear}{t_n}
\newcommand{\timefar}{t_f}

\newcommand{\expo}[1]{\exp\left(#1\right)}

\newcommand{\absrp}{\sigma}

\newcommand{\loss}{\mathcal{L}}
\newcommand{\sampler}{\mathrm{S}}

%% file: math_commands.tex
\usepackage{amsmath,amsfonts,bm}

\def\eqref#1{equation~\ref{#1}}

\def\1{\bm{1}}

\DeclareMathAlphabet{\mathsfit}{\encodingdefault}{\sfdefault}{m}{sl}
\SetMathAlphabet{\mathsfit}{bold}{\encodingdefault}{\sfdefault}{bx}{n}

\DeclareMathOperator*{\argmin}{arg\,min}

%% file: sections/abstract.tex
\begin{abstract}
\normalsize
We present a novel method for performing flexible, 3D-aware image content manipulation while enabling high-quality novel view synthesis. While NeRF-based approaches~\cite{mildenhall2020nerf} are effective for novel view synthesis, such models memorize the radiance for every point in a scene within a neural network. Since these models are scene-specific and lack a 3D scene representation, classical editing such as shape manipulation, or combining scenes is not possible. Hence, editing and combining NeRF-based scenes has not been demonstrated. With the aim of obtaining interpretable and controllable scene representations, our model couples learnt scene-specific feature volumes with a scene agnostic neural rendering network. With this hybrid representation, we decouple neural rendering from scene-specific geometry and appearance. We can generalize to novel scenes by optimizing only the scene-specific 3D feature representation, while keeping the parameters of the rendering network fixed. The rendering function learnt during the initial training stage can thus be easily applied to new scenes, making our approach more flexible. More importantly, since the feature volumes are independent of the rendering model, we can manipulate and combine scenes by editing their corresponding feature volumes. The edited volume can then be plugged into the rendering model to synthesize high-quality novel views. We demonstrate scene manipulation including mixing scenes, deforming objects and inserting objects into scenes, while producing photo-realistic results.
\end{abstract}

%% file: sections/introduction.tex
\section{Introduction}

Scene manipulation and rendering are long-standing problems in graphics, with the goal of creating the desired visual content and providing immersive abilities to explore it. Traditionally, the process consists of acquiring textured meshes of objects and scenes, followed by combining them using specialized software and hardware to reach the desired composition, and finally, rendering the scene using graphical pipelines. Acquiring, composing and rendering are non-trivial problems that require time and experience, and, hence, are not readily available for amateur users.

Impressive progress in Novel View Synthesis (NVS) sparkled by the recently introduced Neural Radiance Fields (NeRF)~\cite{mildenhall2020nerf}, represents an attractive vehicle for scene manipulation and rendering. However, NeRF and most follow-up works suffer from two main shortcomings, limiting their use for creative applications.
First, they require per-scene training, and second, the scene is represented by a neural network, which makes \emph{editing} and manipulation difficult. Recent work has shown how to generalize NeRF to novel scenes~\cite{wang2021ibrnet}, but those works have not demonstrated editing and control. Other recent work have shown editing capabilities by learning per object NeRF models or decomposing a single scene into foreground objects and background~\cite{Schwarz2020NEURIPS,yang2021objectnerf,liu2021editing}. However, these models are either object or scene specific, or work on synthetic scenes without realistic background~\cite{Liu20neurips_sparse_nerf}, limiting its applicability. What is missing is a neural representation and model which allows to represent multiple \emph{real} scenes, while allowing intuitive control. This would retain the realism and simplicity of neural rendering models, while keeping the versatility and intuitive control of traditional computer graphics representations (meshes, volumes and textures).

In this work, we present Control-NeRF, a novel approach which can represent multiple scenes, and allows intuitive control and editing.
Our idea is to decouple the rendering network from the neural scene representation. We learn a latent representation of the scene, encoded as a spatially disentangled feature volume (\ie, in which the point features describe the content and radiance at that point in the scene), coupled with a neural rendering function that computes the radiance and density conditioned on the point feature. 
This decoupled model results in several advantages. 
First, the model can be trained on multiple scenes at once, producing different scene representations for each of them, while learning a general rendering network. 
Second, once the model is learned, new scene representations can be learned 
while holding the rendering network fixed (desirable if for example we want to stream scenes without having to re-train or transmit the rendering network). 
Third, as we show in the experiments, the learned representations are aligned with the real 3D scenes, which allows for intuitive manipulation such as displacing, rotating and displacing objects, integrating objects from other scenes, or simply combining scenes, see Fig.~\ref{fig:teaser}. Most importantly, because each scene has its own representation and the rendering network is shared across scenes, editing and composition can be done post-hoc without re-training. We demonstrate that learning Control-NeRF efficiently on real scenes requires a careful coarse-to-fine strategy --- in which the optimized feature volume dimensions are progressively increased --- and a total variation regularizer on the feature volume representation.
In our evaluations, we demonstrate that our approach allows for NVS using a single model for multiple real training scenes while being comparable to scene specific models.
We also demonstrate how to efficiently generalize to novel scenes by optimizing the scene representation while keeping the rendering network fixed.
Finally, we demonstrate various creative manipulation tasks such as compositing of different real scenes, displacing and rotating objects, and inserting objects. In summary, our primary contributions are:
\begin{itemize}%
\setlength{\parskip}{-1pt}
\setlength{\itemsep}{0pt plus 1pt}
\item A novel NVS model which can learn multiple \emph{real} scene representations at once, while sharing a rendering network. 
\item We demonstrate that the resulting scene representations are aligned with the real 3D scenes allowing for creative 3D-aware manipulation and editing without having to retrain the model.
\item Extensive evaluations demonstrating our approach significantly outperforms competing approaches in terms of the types of manipulations achievable while enabling high-quality image synthesis.
\item We will release our code and models for research purposes.

\end{itemize}

%% file: sections/related_work.tex
\section{Related Work}
\label{sec:related}
\textbf{Novel View Synthesis.} 
Novel-view-synthesis (NVS) is a widely studied problem in the area of image-based rendering. Most NVS methods are focused on warping or blending the input images and inpainting the occluded regions. Recent NVS efforts \cite{riegler2020free,riegler2020stable,aliev2019neural,hedman2018deep} have achieved high quality results relying on geometry proxies, such as  rough reconstructions, depth maps or point-clouds to warp the input images to the target view. Many works use the ability of generative networks to to hallucinate occluded regions from one or a few images~\cite{tatarchenko2015single,sitzmann2019scene,meshry2019neural}, which can be complemented with the use of appearance flow~\cite{zhou2016flow,park2017transformation,sun2018nvs}. However, in case of large viewpoint transformations, best results are attained only for simple/synthetic scenes, or by using a large number of input images. Other techniques, \eg multiplane images~\cite{zhou2018stereo,wizadwongsa2021nex}, have proven suitable for large-scale scenes like those captured in real photographs. Most of these frameworks, however, focus only on NVS and typically provide little or no ability to edit the scene content (\eg, adding or deforming objects).

\textbf{Implicit Surface and Appearance Representations.}
The use of implicit surfaces~\cite{Mescheder19cvpr_occupancy_net,Park19cvpr_deepsdf,Chabra20eccv_DLS,chibane2020implicit} for geometry and appearance reconstruction has proven popular in recent works, with their ability to capture detailed objects with varying topology at arbitrary resolutions. Methods such as PIFu~\cite{Saito19Iccv_PIFu} use these to capture the surface and texture of dynamic humans from monocular images, an approach that was refined and improved in~\cite{saito2020pifuhd,li2020monocular,huang_arch_2020-1} to allow for higher fidelity and realtime performance capture. Methods like these can be used for NVS simply by rendering the obtained reconstructions. However, while they achieve impressive results for individual objects, they struggle to capture the full geometry and appearance of complex real scenes.

\textbf{Volumetric Representations.}
Explicit voxel grids~\cite{kajiya84} have recently been employed for various tasks related to implicit surface and appearance representation, including generative modeling of 3D objects~\cite{abstractionTulsiani17,zhu2018von}, shape and appearance reconstruction from images~\cite{10.1145/2047196.2047270,tulsiani2017ray,sun2018im2avatar,Olszewski_2019_ICCV,Knoche20CVPRW}. Other recent works, such as \cite{sitzmann2019deepvoxels} have explored the use of latent representations with a volumetric structure to implicitly encode a scene's appearance and structure for neural rendering. They use multiple images of static objects to learn a feature volume that can be resampled to a given camera's viewpoint. 
Also, \cite{Lombardi19siggraph_Neural_Volumes} use multiple calibrated images of static and dynamic scenes to learn a latent volumetric representation that can be used for rendering of novel views, including time varying effects (\eg human motion). However, these works do not allow for interactive editing or manipulation of the scene, and are typically scene specific, requiring separate networks for each captured scene. Some methods, such as \cite{Olszewski_2019_ICCV} train a network which infers a latent volumetric representation of previously unseen images that can be spatially transformed to allow for NVS and editing. However, the image quality of the manipulated objects is relatively low, and it only works on simple scenes.

\textbf{Hybrid Latent and Geometric Representations.}
Other recent approaches combine explicit representations of a scene's geometry with a latent representation to exploit neural rendering techniques. Some methods learn neural textures~\cite{thies2019neural} used in conjunction with UV-maps to allow for realistic image synthesis and manipulation. The learnt textures, however, are specific to the corresponding objects and scenes used during training, and thus cannot generalize to new scenes without retraining. In NPBG~\cite{aliev2019neural}, given several images of a scene with a corresponding 3D point cloud, neural descriptors are fitted to points, which are then used with the input data to learn to infer novel views of the scene. This work requires a point cloud of the scene, obtained using multi-view stereo or depth sensor data as part of the training process. Therefore the overall quality of the final results depends heavily on the quality of the reconstruction.

\textbf{Neural Radiance Fields.} Neural Radiance Fields (NeRF)~\cite{mildenhall2020nerf} builds on prior work on implicit surface representations by introducing a sophisticated MLP architecture trained to produce an estimate of the density and outgoing radiance throughout the scene. Volume rendering techniques are used to enforce consistency with the training images, which enables the inference of high-quality novel views of the scene. Subsequent efforts have addressed various limitations of this work and extended it to new applications, \eg accelerating its training and rendering performance and quality~\cite{Tancik20arxiv_meta,Liu20neurips_sparse_nerf,Rebain20arxiv_derf,neff2021donerf,wang2021ibrnet}; extending it to large-scale scenes~\cite{MartinBrualla20arxiv_nerfw,Zhang20arxiv_nerf++}; allowing for the capture and synthesis of dynamic scenes with non-rigid regions, including human heads and bodies~\cite{Park20arxiv_nerfies,Pumarola20arxiv_D_NeRF,Wang20arxiv_hybrid_NeRF}; relighting the captured content~\cite{Bi20arxiv_neural_reflection_fields,Boss20arxiv_NeRD,Srinivasan20arxiv_NeRV}; camera and body pose estimation~\cite{Su21arxiv_A_NeRF,YenChen20arxiv_iNeRF}; and NVS with unknown camera parameters \cite{Wang21arxiv_nerfmm}.
Some works \cite{Trevithick20arxiv_GRF,Yu20arxiv_pixelNeRF,Chibane_2021_CVPR} use projected features from images into a space that may be queried in a manner similar to~\cite{Saito19Iccv_PIFu,li2020monocular}. With a NeRF-like radiance function they demonstrate the ability to perform NVS using a single or few input images. However the overall quality and complexity of the synthesized images is limited, and they do not enable general manipulations of the scene, as in our method.
Recently there have been few works that combine voxel grids and neural radiance fields. \cite{Liu20neurips_sparse_nerf} use sparse voxel fields to learn local radiance fields for improved rendering performance. For the given scene they build the voxels by pruning the voxel grid at training time. They can also do local shape editing and build scenes by compositing separate objects together. While this method shows impressive results on individual objects, they struggle to deal with real scenes with complex background and front facing scenes, where the scene is not observed from all sides. Another similar work, \cite{yang2021learning} has introduced a method that learns an object-compositional neural radiance field. They learn separately a scene branch to encode the scene appearance and individual object branches for all the object in the scene. This method allows for object-level editing, such as moving and transforming the objects in the scene. However unlike our method it is scene-specific and does not support moving objects across multiple scenes. For a more comprehensive survey of work in this area, please refer to~\cite{dellaert2021neural}.

%% file: sections/method.tex
\section{Control-NeRF}
\label{sec:method}

\begin{figure*}
  \centering
  \includegraphics[width=1.00\linewidth]{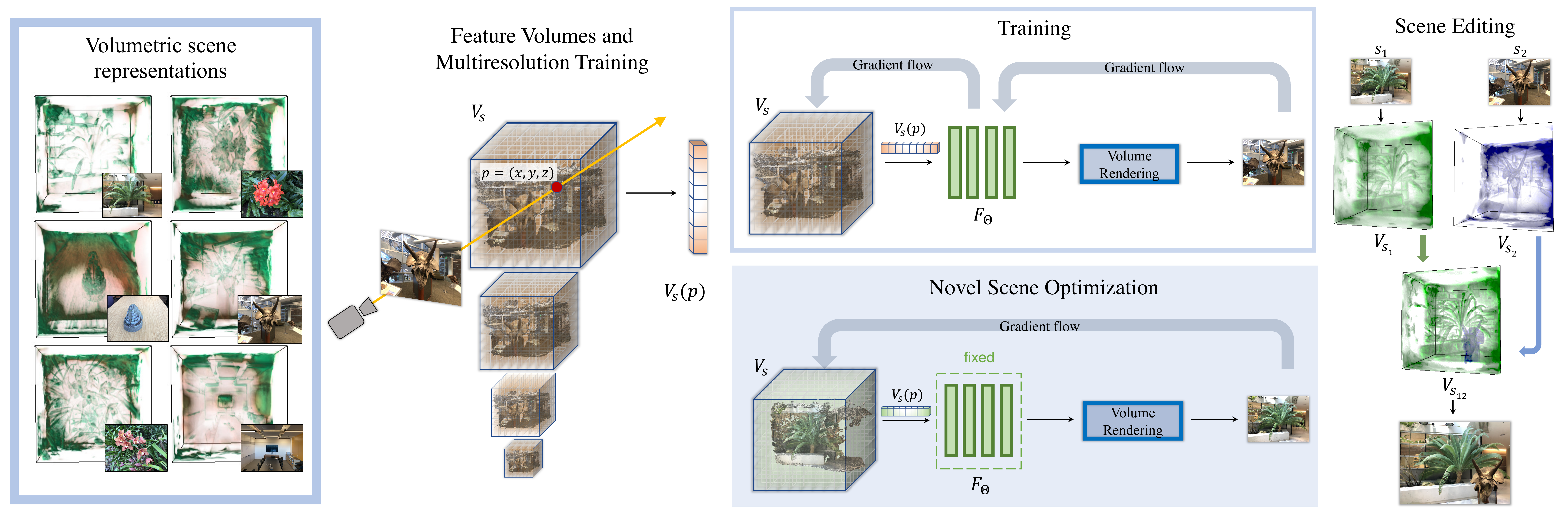}
      \caption{Our method learns a volumetric representations for multiple scenes simultaneously. Left in the figure we show visualizations of the learned feature volumes. We query the volume along the ray and predict color and density based on the obtained features. The pixel color is derived using volume rendering, similar to~\cite{mildenhall2020nerf}. At training time the volume and the rendering network are trained jointly. For novel scenes, the rendering network is fixed and only the scene volume is optimized. As shown on the right, these volumes can be edited and mixed and for the purpose of scene editing.}
  \label{fig:pipeline}
\end{figure*}

We present our novel-view synthesis method, Control-NeRF, (Figure~\ref{fig:pipeline}) that is based on feature volumes and Neural Radiance Fields \cite{mildenhall2020nerf} and allows for scene editing, mixing and manipulation. We decouple geometry/appearance from rendering by learning dense feature volume as representation for every scene and a single rendering model that generalizes across scenes.  The rendering model takes a feature vector sampled from the volume and predicts density and color value. As shown in~\cite{mildenhall2020nerf}, these predictions are used as input to a volume rendering function that accumulates the point along a ray to generate a pixel color. 

This section of the paper is organized as follows. In Sec.~\ref{sec:nerf}, we briefly review the general framework used for performing novel view synthesis using neural radiance fields.
Sec.~\ref{sec:vol_rep} describes how we make use of learnt feature volumes to condition the radiance field output on a given scene.
In Sec.~\ref{sec:train_net}, we describe the losses and the training procedure for optimizing the network parameters and per-scene feature volumes. We also show how to learn feature volumes for novel scenes not seen at training time. (Sec.~\ref{sec:train_gen}).
Sec.~\ref{sec:scene_edit} describes how we can use the learnt feature volumes for arbitrary creative scene manipulations and render the result.

\subsection{Background}
\label{sec:nerf}
Most works based on Neural Radiance Fields~\cite{mildenhall2020nerf}
predict radiance and color for a pair of point and viewing angle direction of a single scene:

\begin{equation}
F_{\modeltheta} : (\gamma(\mathbf{p}), \gamma(\mathbf{d})) \to (\mathbf{c},\sigma)
\label{eq:nerf}
\end{equation}

where $\mathbf{c} \in \mathbb{R}^3$ is an RGB value indicating the radiance from point $\mathbf{p} \in \mathbb{R}^3$ in direction $\mathbf{d} \in \mathbb{R}^3$, and $\sigma \in \mathbb{R}$ is the density value $\mathbf{p}$, indicating how much the radiance contributes to view rays intersecting the scene at that point.
Optionally, one can use $\gamma$ which is a positional encoding~\cite{NIPS2017_3f5ee243} used to allow this network to better capture high-frequency details.
Images are rendered one pixel at a time, using volumetric sampling techniques, querying the MLP at points along the camera ray $\mathbf{r}(t) = \mathbf{o} + t\mathbf{d}$ (where $\mathbf{o}$ indicates the camera origin and $t$ indicates the distance from the origin along the ray) corresponding to that pixel. By integrating the radiance values at a point using its density, the appropriate color values can be computed. 

The problem with NeRF based approaches is that the scene is memorized within the neural network, which makes compositing of scenes and editing hard. 
 
\subsection{Formulation}
\label{sec:vol_rep}
To allow realistic editing, our method decouples the scene representation from the neural rendering network. Instead of memorizing a mapping from scene point and viewing directions to radiance with an MLP as in Eq.~\ref{eq:nerf}, we learn a scene-specific volume of deep features. Then a rendering network maps from deep features extracted at continuous locations of the volume, to radiance and color. 
\paragraph{Scene Representation} 
Given a set of input RGB images $\mathcal{I} = \{I_s^i\}_{i=1}^{N_s}$ from $M$ training scenes $s \in \mathcal{S}$, $M=|\mathcal{S}|$, we seek to learn a latent volumetric representation $V_s \in \mathbb{R}^{WHDF}$ for each scene $s$, with a spatial resolution of $W \times H \times D$ and a feature vector of length $F$ in each cell, which can be both rendered from novel views and edited to allow for novel manipulations of the scene content while still allowing for high-quality view synthesis. We use a resolution of $W=H=D=128$ and a feature vector of length $F = 64$ in our experiments.
\paragraph{Rendering Network}

The rendering network is a learned mapping from a deep feature $\mathbf{v}_s \in \mathbb{R}^{64}$ to radiance and color. 
The deep feature describes the local shape and appearance of the corresponding position $\mathbf{p} = (x, y, z)$ in scene $s$ extracted from a scene-specific volume of deep features.  Mathematically, 
\begin{equation}
F_{\modeltheta} : (\sampler(V_s, \mathbf{p})), \gamma(\mathbf{d})) \to (\mathbf{c},\sigma)
\label{eq:control-nerf}
\end{equation}
where the feature vector $\mathbf{v}_s$ is obtained by sampling of the feature volume $\mathbf{v}_s = \sampler(V_s, \mathbf{p})$, where $\sampler$ indicates the trilinear resampling operation. As in NeRF the density $F_{\modeltheta}$ is integrated along rays $\ray$ to produce pixel colors -- this operation is denoted by ${\hat{C}(\ray,V_s,\modeltheta)}$. 
In contrast to ~\eqref{eq:nerf}, the formulation in ~\eqref{eq:control-nerf} allows us to optimize the volume $V_s$ for each scene, while simultaneously learning the parameters of the density network $F_{\modeltheta} : (\mathbf{v}_s, \gamma(\mathbf{d})) \to (\mathbf{c},\sigma)$. After this initial training stage, the parameters $\modeltheta$ of this rendering module are fixed. For every novel scene, we only optimize its feature volume $V_s$. 
This will allow us to combine and edit scenes by manipulating their respective feature volumes $V_s$, and render the result using the general rendering module ${\hat{C}}$. 

Note: As in NeRF, in practice, 2 networks are trained: a coarse network in which samples are taken from evenly-spaced intervals along the view ray, and a fine network  which uses the density values from the coarse network to select sample points more likely to contribute to the corresponding ground-truth pixel color value $\Ctrue$. 
In the following, we denote the density integrals of the coarse and fine networks as $\Cfine$,
$\Ccoarse$ respectively. ~\footnote{For simplicity, from here on we use $\modeltheta$ to refer to the total learnable parameters of the 2 rendering networks.}

\subsection{Training Strategy and Generalization}
\label{sec:train_net}

\subsubsection{Training Losses}
\label{sec:train_loss}
\paragraph{Reconstruction Loss.}
Our primary loss is a straightforward reconstruction loss on the rendered pixel values. As in~\cite{mildenhall2020nerf}, at each iteration we randomly sample and integrate a subset of the rays $\raybatch_s$ from the images for the current scene $s$, and compute the mean-squared error between them and the corresponding pixels in the ground-truth images:

\begin{equation}
\label{eq:reconloss}
  \begin{aligned}
    &\Ltrain_r(\raybatch_s,\trainsceneimages_s,V_s,\modeltheta)=\\
    &\mathbb{E}_{\ray\sim\raybatch_s}\left[\norm{\Ccoarsefull-\Ctrue}_2^2+\norm{\Cfinefull-\Ctrue}_2^2\right]
  \end{aligned}
\end{equation}

Using this loss for each training scene (see Sec.~\ref{sec:multi_training}), we jointly optimize the network parameters $\modeltheta$ and the feature volumes $\trainscenevolumes$ for all training scenes.
\paragraph{Total Variation Loss.} One very useful property that we want the volumes to exhibit is that neighbouring feature vectors should have similar values. NeRF~\cite{mildenhall2020nerf}, has this property by default, since it relies on $\mathbb{R}^3$ (3D locations as input). In order to encourage similar behaviour for our feature volumes we add regularization. In our experiments, we found that a more consistent and coherent feature volume was learned if we introduce a total variation regularization~\cite{Rudin92nonlineartotal} loss to the learned feature volume. 
To reduce the memory usage and computation introduced by this loss, we apply it on the $64$-dimensional feature vectors in a randomly sampled contiguous subregion $R \subset V_s$ that is $1/4$ of the current latent feature volume dimensions (see Sec.~\ref{sec:hier_training}) for the current scene $s$ during each training iteration.

\begin{equation}
  \label{eq:totalvarloss}
  \loss_{tv}(V_s)=\mathbb{E}_{R \sim V_s}\left[\left|
  T(R)
  \right|\right],
\end{equation}

\begin{equation}
  T(R) = \sum_{i, j, k}
  \sqrt{
    \begin{aligned}
      |R_{i+1, j, k}-R_{i, j, k}|^{2} & +|R_{i, j+1, k}-R_{i, j, k}|^{2} \\
      + |R_{i, j, k+1} & -R_{i, j, k}|^{2}
    \end{aligned}
  }
\end{equation}

Thus, we minimize the following total loss function by optimizing the parameters $\modeltheta$ and feature volumes $\trainscenevolumes$ corresponding to the calibrated images $\trainsceneimages$ and the corresponding view rays $\raybatch$ for each training scene:

\begin{equation}
  \begin{aligned}
    & \argmin_{\trainscenevolumes,\modeltheta}\Ltrain(\raybatch,\trainsceneimages,\trainscenevolumes,\modeltheta)=\\
    & \mathbb{E}_{s \sim \mathcal{S}}\left[\Ltrain_r(\raybatch_s,\trainsceneimages_s,V_s,\modeltheta)+\lambda\loss_{tv}(V_s)\right]%
  \end{aligned}
\end{equation}

where $\lambda=10^{-4}$.

\subsubsection{Multi-Resolution Volume Training}
\label{sec:hier_training}
As the final volume contains a $64$-dimensional feature vector per cell in the $128^3$ volume, training the network at this full resolution is quite intensive. As such, we employ a hierarchical training process to compute these volumes in a coarse-to-fine manner.
This allows for improved training time while retaining the ability to perform high-quality image synthesis and manipulation.
We start training with a feature volume resolution of $16^3$.
The model is trained until convergence, optimizing both the current feature volume $V_s$ and rendering module parameters $\modeltheta$.
We then upsample the learnt feature volume to increase its dimensions by a factor of 2, and proceed to train until convergence at the new resolution. We use $4$ stages in our hierarchical training process, doubling the feature volume dimensions at each stage until we reach the target resolution of $128^3$. 
\subsubsection{Multi-Scene Training}
\label{sec:multi_training}
To allow the rendering module to be employed for multiple scenes, it needs to be trained in a multiscene scenario.
During training we randomly select one of the scenes $s \in \mathcal{S}$ and load its feature volume $V_s$, then train using rays sampled from this volume for several consecutive iterations, before saving the feature volume and repeating the process with a new randomly selected scene.
While sampling a new scene at each training iteration would better approximate the effect of incorporating samples from multiple scenes at each step in the optimization, this would require additional overhead as feature volumes are loaded into GPU memory, then copied back to be stored for their next use.
We empirically found that $50$ consecutive iterations between scene transitions produced a sufficient balance between training performance and multi-scene representation capacity.

\subsubsection{Generalization to Novel Scenes}
\label{sec:train_gen}
After the initial training stage in which the parameters $\modeltheta$ of the radiance network $F_\modeltheta$ are trained in conjunction with the optimization of the $M$ per-scene feature volumes $V_{1,...M}$, we allow for efficient generalization to novel scenes by fixing the parameters $\modeltheta$ and solely optimizing the parameters of the feature volumes corresponding to these novel scenes.

Given a new set of scenes $\testsceneimages$ not used during the initial training stage, and a set of images corresponding to each scene $\mathcal{I}^{'} = \{I_g^i\}_{i=1}^{N_g}$ for each scene $g \in \testsceneimages$, we perform the optimization process as described above, while \emph{only} optimizing the corresponding feature volume $V_g^{'}$ for each scene.
We employ the hierarchical training strategy defined in Sec.~\ref{sec:hier_training}
, and the losses defined in Eqns.~\ref{eq:reconloss} and~\ref{eq:totalvarloss}, but for these scenes only optimize the feature volumes corresponding to each scene $g$ to minimize the total loss:

\begin{equation}
  \begin{aligned}
    & \argmin_{\trainscenevolumes^{'}}\Ltrain(\raybatch^{'},\trainsceneimages^{'},\trainscenevolumes^{'},\modeltheta)=\\
    & \mathbb{E}_{g \sim \mathcal{G}}\left[\Ltrain_r(\raybatch_g^{'},\trainsceneimages_g^{'},V_g^{'},\modeltheta)+\lambda\loss_{tv}(V_g^{'})\right]%
  \end{aligned}
\end{equation}

Given sufficient training scenes, the learnt radiance function can be applied to optimize for novel scenes more efficiently than when training to infer the volumes and network parameters together as in the initial training process. In our experiments we show that a small number of training scenes (we used only 6) are sufficient to train a generalizable radiance function. 

\subsection{Scene Editing and Manipulation}
\label{sec:scene_edit}
Our volumetric representation of scene-specific content allows for scene manipulations by manipulating it's feature volume. By applying the trilinear resampling operation $\sampler$ defined in Sec.~\ref{sec:vol_rep} to contiguous subregions of the feature volume (or the entire volume, if global scene deformations are desired), 
nonrigid spatial manipulations can be applied.
If $V_o$ is the original feature volume and $P \in \mathbb{R}^{3WHD}$ is a matrix of 3D coordinates indicating where to sample from for each point in the modified volume (which may be the coordinates of the corresponding point in the original volume, for stationary regions), $V_m = \sampler(V_o, P)$ will produce a volume with the desired spatial deformation. Even more so, since our radiance function can generalize across scenes, features from multiple scenes can be mixed together to simulate inserting objects from one scene into another.

%% file: sections/experiments.tex
\begin{figure*}[ht]
  \begin{center}
    \includegraphics[width=1.0\textwidth]{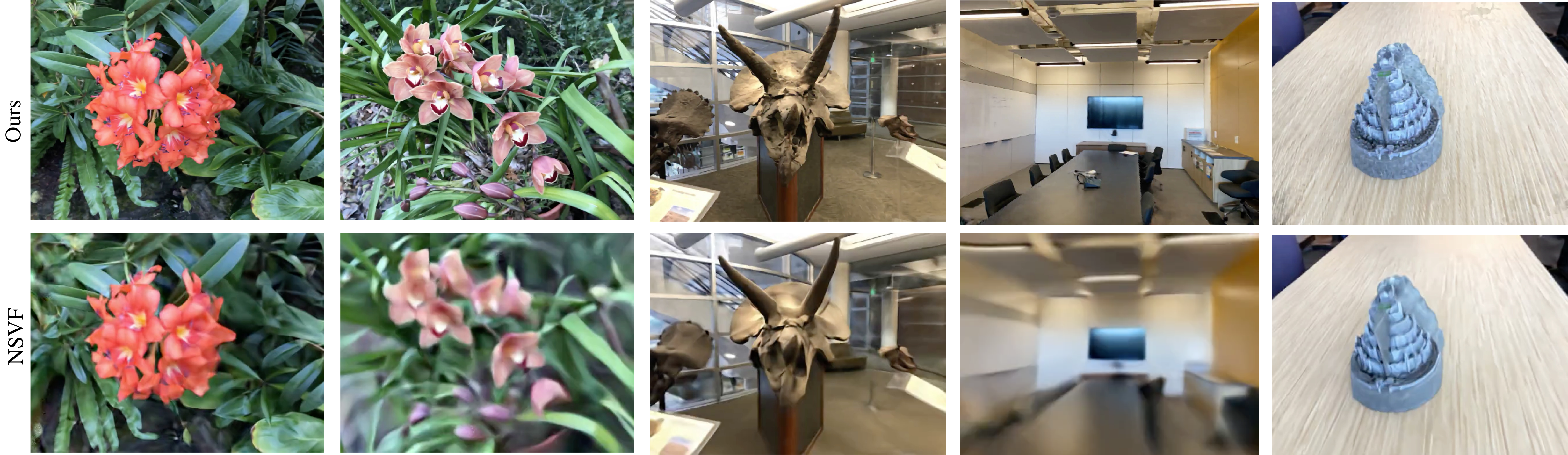}
  \end{center}
  \captionof{figure}{\textbf{Comparison to NSVF~\cite{Liu20neurips_sparse_nerf}} in Novel view synthesis. As discussed, NSVF struggles with real frontal scenes, in which the content is not captured from 360$^{\circ}$.}
  \label{fig:nsvf_comparison}
\end{figure*}

\begin{figure*}[ht]
  \begin{center}
    \includegraphics[width=1.0\textwidth]{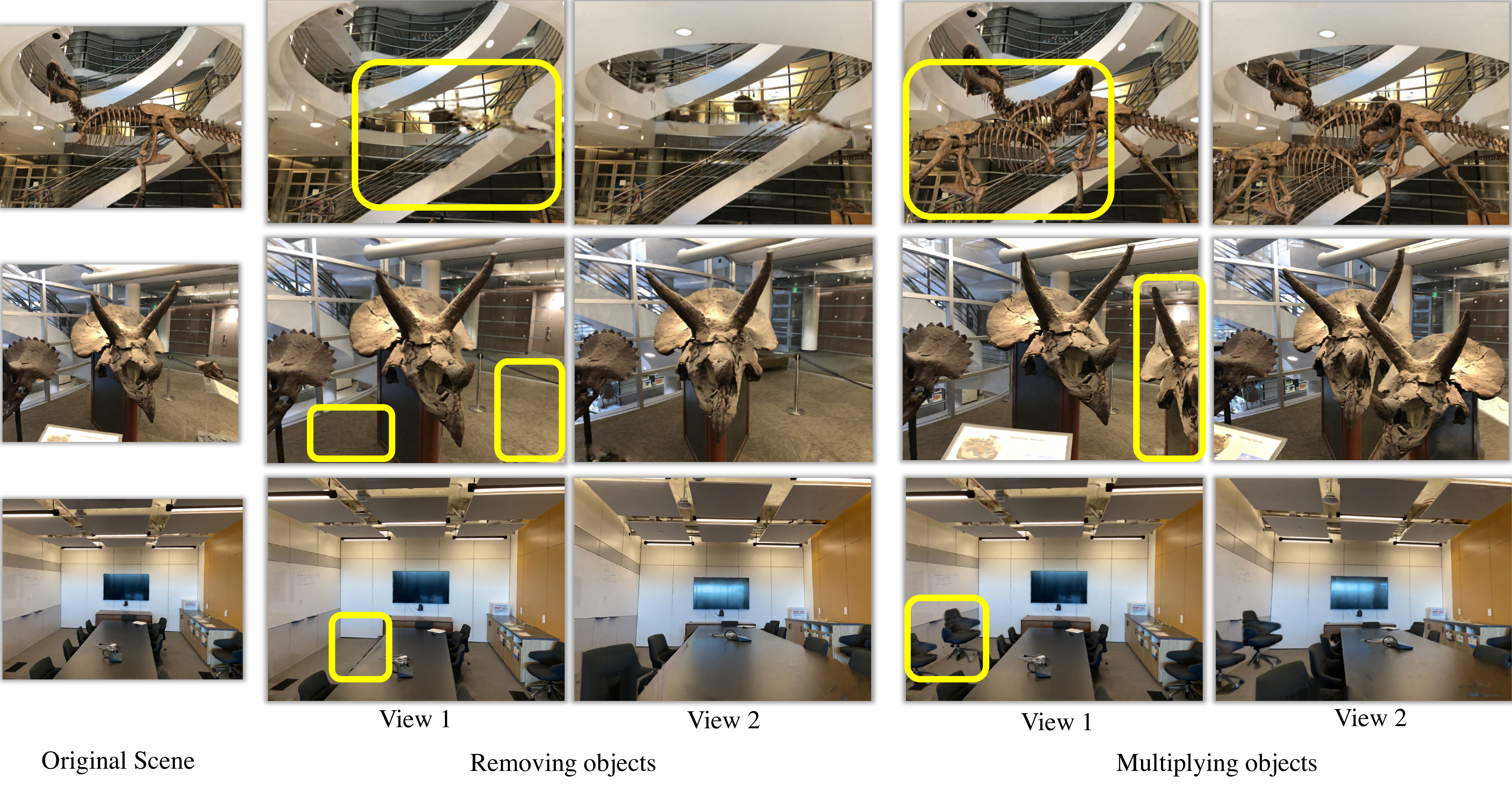}
  \end{center}
  \captionof{figure}{\textbf{Replicating and removing object from scenes} The first column shows the original scene. The rest of the columns show the edited scene from two different views. The differences are marked with yellow rectangles in the first view.}
  \label{fig:add_remove}
\end{figure*}

\begin{figure*}[ht]
  \centering
  \includegraphics[width=1.0\linewidth]{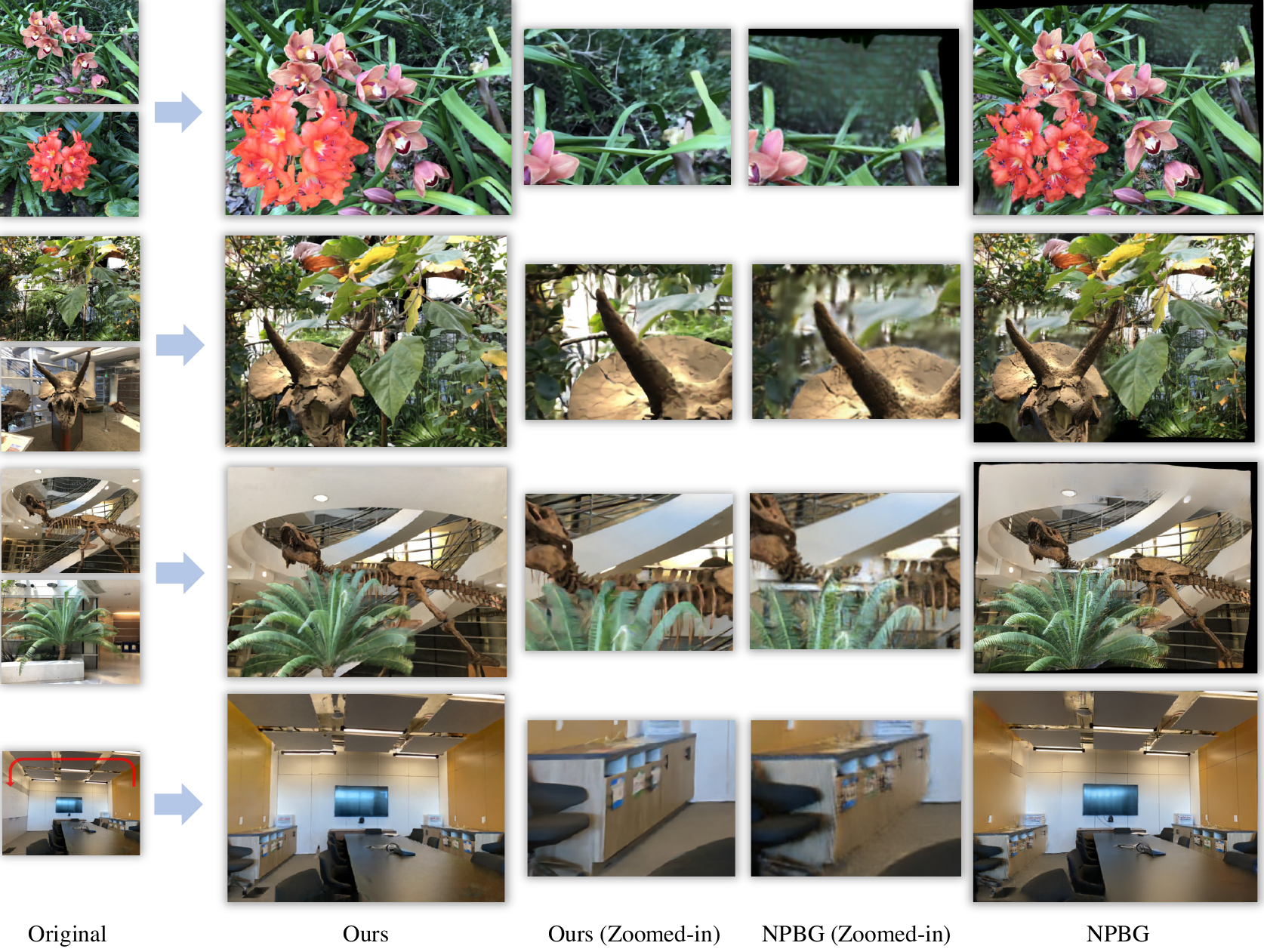}
  \captionof{figure}{\textbf{Scene Editing Results.} The first column shows the original scenes that are used in the edit. An object from the source scene is replicated in the target scene. The following columns show novel view of our editing results and results obtained using NPBG~\cite{aliev2019neural}. We kindly ask readers to zoom in on these images and consult the supplementary video for more results and animations.
  }
  \label{fig:edited}
\end{figure*}

\section{Experiments and Results}

\subsection{Dataset and Implementation Details}
\label{sec:expr_impl}
For our initial training stage, we use $6$ scenes from the dataset provided by LLFF~\cite{mildenhall2019llff}, consisting of a total of 230 images (an average of approximately 38 images per scene) with the cameras' extrinsic and intrinsic parameters estimated by COLMAP~\cite{colmap}.
After this stage, we fix the rendering module parameters and optimize the feature volumes for new scenes individually.
We use $2$ scenes from this dataset, withheld during the initial training stage, to demonstrate our novel scene generalization capabilities (\emph{fern} and \emph{trex}, shown in Figure~\ref{fig:teaser}), consisting of a total of 75 images. Please consult the appendix for more details.

\subsection{Scene Content Manipulation}
\label{sec:expr_scene_edit}
Using the scene resampling and editing techniques described in Sec.~\ref{sec:scene_edit}, we demonstrate various creative manipulations enabled by our method. In Figure~\ref{fig:edited} we show scene manipulation by moving object from one scene into another. The scenes shown in these examples are real scenes from the LLFF \cite{mildenhall2019llff} dataset. 

In Figure~\ref{fig:add_remove}, we show single scene editing by removing objects or making copies of existing objects.

\subsection{Evaluations and Comparisons.}
\label{sec:expr_compare}

\paragraph{Editing Comparisons.}
We provide qualitative and quantitative comparisons of our approach to scene manipulation to two related methods, Neural Point-Based Graphics (NPBG)~\cite{aliev2019neural} and Neural Sparse Voxel Fields (NSVF)~\cite{Liu20neurips_sparse_nerf}. NPBG uses 3D point clouds of a scene with corresponding RGB images and camera poses to allow for both realistic neural rendering of novel views of the scene and copying content from one scene into another. NSVF uses sparse voxels scene representation that is pruned at training time. This representation is useful for isolated objects, but struggles with real scenes with complex background. 

\begin{table}
\begin{center}
\footnotesize
\tabcolsep=0.11cm
\begin{tabular}{|l||c|c|c|} \hline 
 & PSNR $\uparrow$ & SSIM $\uparrow$ & LPIPS $\downarrow$ \\ \hline \hline
NSVF~\cite{Liu20neurips_sparse_nerf} & 20.414 & 0.536 & 0.449 \\ \hline 
NPBG~\cite{aliev2019neural} & 19.430 & 0.727 & 0.242 \\ \hline 
Ours & \textbf{25.635} & \textbf{0.853} & \textbf{0.181} \\ \hline 
\end{tabular}
\end{center}
\caption{\textbf{Quantitative comparison with NPBG~\cite{aliev2019neural} and NSVF~\cite{Liu20neurips_sparse_nerf}}. Metrics are computed across test images for scenes from from LLFF~\cite{mildenhall2019llff} dataset. ``Ours'' is our method trained on 6 scenes simultaneously as in our original setup. Please consult the appendix for more detailed comparisons.}
\label{tab:eval_npgb_nsvf}
\end{table}
For our quantitative comparisons we evaluate the novel-view-synthesis capabilities of our method in comparison to NPBG and NSVF (Table~\ref{tab:eval_npgb_nsvf}). For the qualitative comparisons we evaluate the scene editing capabilities of our method in comparison to NPBG (Figure~\ref{fig:edited}). While in theory NSVF could perform similar manipulations, the official implementation doesn't support multi-scene editing or scene manipulation. Nevertheless in Figure~\ref{fig:nsvf_comparison} we compare our method to NSVF in the task of novel view synthesis of complex real scenes.
We could not compare to the recent method of \cite{yang2021learning}, as there is currently no released implementation.
Using $43$ images from the $8$ aforementioned scenes
(The $6$ initial training scenes and the $2$ scenes optimized with fixed renderer parameters)
withheld during the training process, we compute the Peak Signal-to-Noise Ratio (PSNR), Structural Similarity Index (SSIM), and Learned Perceptual Image Patch Similarity (LPIPS)~\cite{zhang2018unreasonable} between the ground truth and the images synthesized using both methods.
Table~\ref{tab:eval_npgb_nsvf} contains the results, which show that our method outperforms~\cite{aliev2019neural} and ~\cite{Liu20neurips_sparse_nerf} using each metric.
In Fig.~\ref{fig:edited} we show that, while both approaches can be used to combine scenes, our approach outperforms~\cite{aliev2019neural} when it comes to editing capabilities.

\paragraph{Additional Results and Details.}
Please consult the supplementary video and document for animated results from these experiments, as well as further results and details on our approach and evaluations.
We include additional NVS and manipulation results for various datasets.
In all these scenarios, we only train the feature volumes of the new scenes while keeping the rendering parameters fixed.
We also provide the results of experiments with non-rigid scene manipulation and articulated animation; an ablation study evaluating the utility of our approaches key components; a perceptual study asking users to evaluate our approach compared to NPBG~\cite{aliev2019neural}; and comparisons with the original NeRF~\cite{mildenhall2020nerf}, as well as a modified version of this work adapted to handle multiple scenes.

%% file: sections/conclusion.tex
\section{Conclusion}

With our method we have explored a promising method for flexible scene manipulation attainable with neural radiance fields. In disentangling between rendering and scene representation, our approach extends the work of state-of-the-art NVS methods to enable practical techniques for efficient multi-scene training and high-fidelity image synthesis and manipulation. We have shown a wide range of possible edits, such as replicating and removing objects, applying rigid and non-rigid transformations and moving objects from one scene to another.

In future work, we intend to explore methods to enhance our approach with more editing options: modifying the textures and appearance of scene content; producing more realistic edits by adapting to different lighting conditions, \eg shadows; and exploring techniques for more efficient scene optimization and rendering.

\vspace{0.5em}

\small
\noindent {\bf Acknowledgements:} We thank Aymen Mir, Bharat Bhatnagar, Garvita Tiwari, Ilya Petrov, Jan Eric Lenssen, Julian Chibane, Keyang Zhou and Xiaohan Zhang for the in-depth discussions, valuable insights and honest feedback. 

This work is and supported by the German Federal Ministry of Education and Research (BMBF): Tübingen AI Center, FKZ: 01IS18039A; and partially funded by the Deutsche Forschungsgemeinschaft (DFG, German Research Foundation) - 409792180 (Emmy Noether Programme, project: Real Virtual Humans).

Gerard Pons-Moll is a member of the Machine Learning Cluster of Excellence, EXC number 2064/1 – Project number 390727645.

%% file: supp_sections/training.tex
\vspace{-1.0em}
\section{Training and Architecture Details}
\label{sec:supp_training}
\vspace{-0.5em}

As noted in Sec. 3.1, we use a positional encoding function~\cite{NIPS2017_3f5ee243} on each dimension of the direction vector $\mathbf{d} \in \mathbb{R}^3$, to map this vector into a higher-dimensional space before passing it as input to our radiance function $F_{\modeltheta} : (\mathbf{v}_s, \gamma(\mathbf{d})) \to (\mathbf{c},\sigma)$, where

\newcommand{\dummypos}{p}
\newcommand{\numfrequencies}{L}
\vspace{-0.5em}

\begin{equation}
    \gamma(\dummypos) = \left(
    \begin{array}{ccccc}
    \sin\left(2^0 \pi \dummypos\right),
    \cos\left(2^0 \pi \dummypos\right),
    \cdots,\\
    \sin\left(2^{\numfrequencies-1} \pi \dummypos\right),
    \cos\left(2^{\numfrequencies-1} \pi \dummypos\right)
    \end{array} \right).
\label{eq:enc}
\end{equation}

We use $L = 4$ in our experiments.
\vspace{-0.5em}

\begin{equation}
\Ctrue = \int_{\timenear}^{\timefar}T(t)\absrp(\mathbf{v}_s)\mathbf{c}(\mathbf{v}_s,\mathbf{d})dt\,,
\end{equation}

\begin{equation}
T(t) = \expo{-\int_{\timenear}^{t}\absrp(\mathbf{v}_s)ds}
\end{equation}

where $\mathbf{v}_s = \sampler(V_s, \mathbf{p})$ is the $64$-dimensional sampled feature vector from the volume $V_s$ for scene $s$ at point $\mathbf{p} \in \mathbb{R}^3$, and $\sampler$ represents the trilinear sampling operation.
The density values sampled from the network can thus be used to determine the probability of a ray terminating at the sampled point along the ray.
In practice, following the example of~\cite{mildenhall2020nerf}, we use a discretized approximation of this integral, using a 2-stage process in which we optimize a coarse network $\Ccoarse$ that samples $64$ points from evenly spaced bins along the ray length, followed by sampling these points plus another $64$ points from our fine network $\Cfine$ using the coarse network opacity results to sample from more relevant portions of the scene volume (see Sec. $5.2$ of~\cite{mildenhall2020nerf}.
For our experiments, the networks are trained using $1024$ rays per batch sampled from the LLFF~\cite{mildenhall2019llff} multi-view image datasets, scaled to a resolution of $504 \times 378$.~\footnote{This differs slightly from the training parameters used in~\cite{mildenhall2020nerf}, as they use $4096$ rays per batch sampled from $1008 \times 752$ images and an additional $128$ samples from the fine network for training. See Sec.~\ref{sec:supp_comparisons} of this document for comparisons to the results obtained when training both their and our network using the parameters described above.}
The network architecture we use is overall based on that of~\cite{mildenhall2020nerf}, except that the input channels have been modified to accept our feature vector in place of the parameters representing the point to be sampled in the training scene.
While they use a positionally encoded representation of each dimension in the the 3D position $\mathbf{p}$ sampled along the view ray (with $L=10$, for a total of $60$ parameters passed as input to represent this position in the scene as in Eq.~\ref{eq:enc}), we pass the $64$-dimensional feature vector sampled from the volume as described above into the network with no positiona encoding. 

%% file: supp_sections/ablations.tex
\vspace{-0.5em}
\section{Ablation}
\label{sec:supp_ablation}
\vspace{-0.5em}

We performed an ablation study to evaluate the efficacy of the total variation loss and multi-resolution training techniques described earlier.
We use the trained model to optimize the feature volumes for the \emph{ferns} and \emph{trex} scenes with and without the aforementioned techniques.
We provide the per-scene results of these experiments both with and without the Total Variation loss and multi-resolution training described in Secs. $3.3.2$ and $3.3.3$, using the Peak Signal-to-Noise Ratio (PSNR), Structural Similarity Index (SSIM), and Learned Perceptual Image Patch Similarity (LPIPS)~\cite{zhang2018unreasonable} metrics.
As Table~\ref{tab:ablate} shows, these results show that overall our final approach outperforms these less sophisticated alternatives per-scene in nearly all cases for each metric, and on average for all metrics.
\vspace{-0.5em}

\begin{table}[ht]
\begin{center}
    \footnotesize
    \tabcolsep=0.25cm
    \begin{tabular}{l|c|c||c} %
      \multicolumn{4}{c}{PSNR$\uparrow$} \\
      & Fern & T-Rex & Avg. \\ \hline
      Our method&\textbf{25.752}&\textbf{26.510}& \textbf{26.131} \\ \hline
      w\textbackslash o Multiscale&24.789&25.617& 25.203 \\ \hline
      w\textbackslash o Total Variation (TV)&25.039&25.504& 25.271 \\ \hline
      w\textbackslash o Multiscale, w\textbackslash o TV&22.193&19.296&20.745 \\ \hline
    \end{tabular}
\end{center}
\begin{center}
    \footnotesize
    \tabcolsep=0.25cm
    \begin{tabular}{l|c|c||c} %
      \multicolumn{4}{c}{SSIM$\uparrow$} \\
      & Fern & T-Rex & Avg. \\ \hline
      Our method&\textbf{0.820}&\textbf{0.907}&\textbf{0.864} \\ \hline
      w\textbackslash o Multiscale&0.793&0.878&0.835 \\ \hline
      w\textbackslash o Total Variation (TV)&0.804&0.869&0.836 \\ \hline
      w\textbackslash o Multiscale, w\textbackslash o TV&0.704&0.691&0.698 \\ \hline
    \end{tabular}
\end{center}
\begin{center}
    \footnotesize
    \tabcolsep=0.25cm
    \begin{tabular}{l|c|c||c} %
      \multicolumn{4}{c}{LPIPS$\downarrow$} \\
      & Fern & T-Rex & Avg. \\ \hline
      Our method&0.236&\textbf{0.153}&\textbf{0.195} \\ \hline
      w\textbackslash o Multiscale&0.279&0.209&0.244 \\ \hline
      w\textbackslash o Total Variation (TV)&\textbf{0.230}&0.221&0.226 \\ \hline
      w\textbackslash o Multiscale, w\textbackslash o TV&0.324&0.366&0.345 \\ \hline
    \end{tabular}
\end{center}
\vspace{-1.0em}
\caption{\textbf{Per-scene quantitative ablation results}.}

\label{tab:ablate}
\end{table}

\begin{table}
\begin{center}
    \footnotesize
    \tabcolsep=0.25cm
    \begin{tabular}{l|c|c||c} %
      Num. Scenes & PSNR$\uparrow$ & SSIM$\uparrow$ & LPIPS$\downarrow$ \\ \hline
      1 & 24.1 & 0.79 & 0.29 \\ \hline
      3 & 25.13 & 0.83 & 0.26 \\ \hline
      6 & \textbf{26.13} & \textbf{0.86} & \textbf{0.19} \\ \hline
    \end{tabular}
\end{center}
\vspace{-1.0em}
\caption{\textbf{Quantitative results on generalization w/ varying number of training scenes}.}
\vspace{-1.0em}
\label{tab:scene_ablate}
\end{table}

%% file: supp_sections/comparisons.tex
\section{Additional Evaluations}
\label{sec:supp_comparisons}

\subsection{Generalization to Novel Content}

\begin{figure}
  \centering
  \includegraphics[width=\columnwidth]{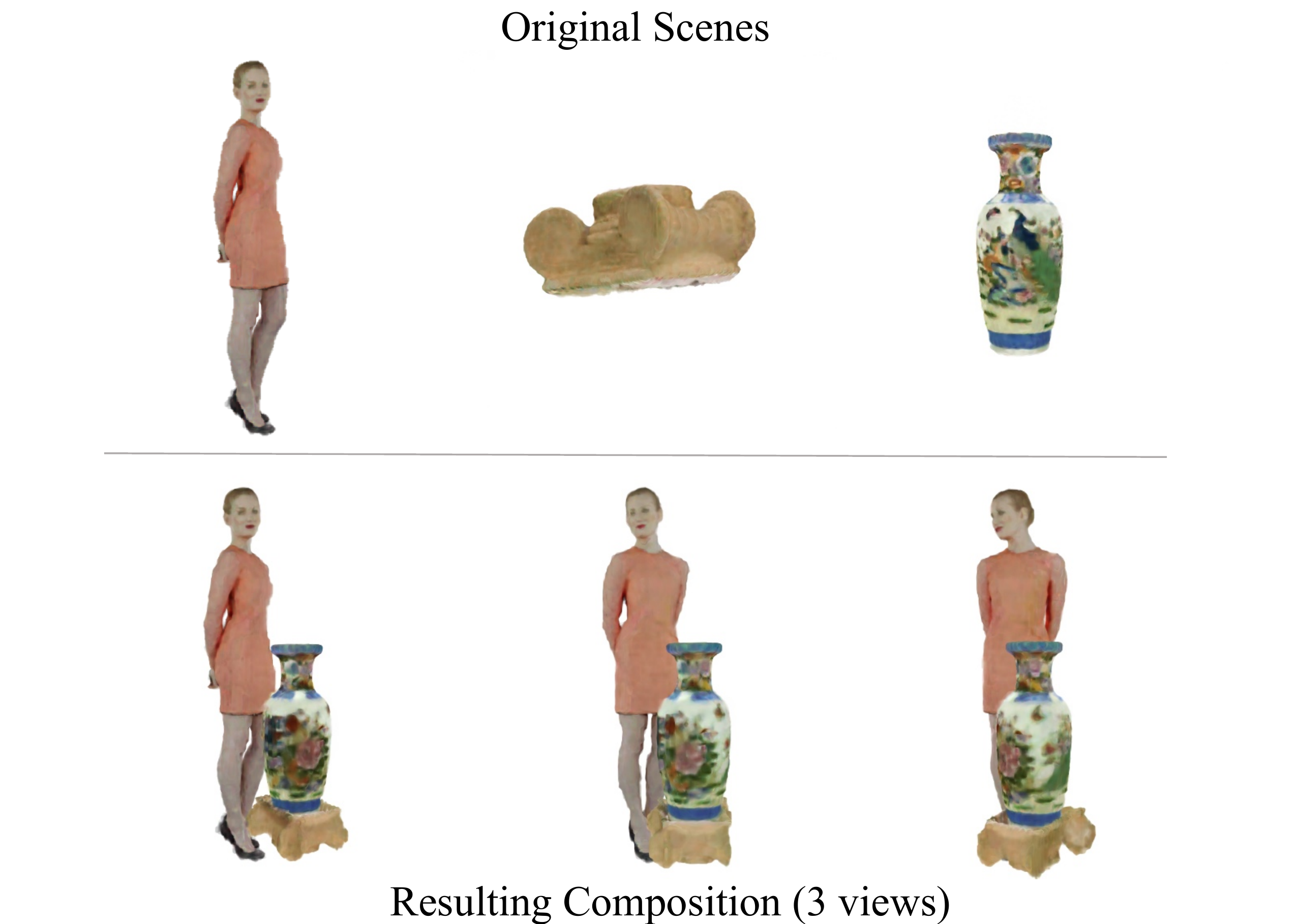}
  \captionof{figure}{\textbf{Combining objects from various datasets}. Note that we use the rendering network trained on 6 scenes from LLFF~\cite{mildenhall2019llff} and finetune a volume for each object separately.
  }
  \vspace{-1.5em}
  \label{fig:composition}
\end{figure}

\begin{figure}
  \vspace{-1.5em}
  \centering
  \includegraphics[width=\columnwidth]{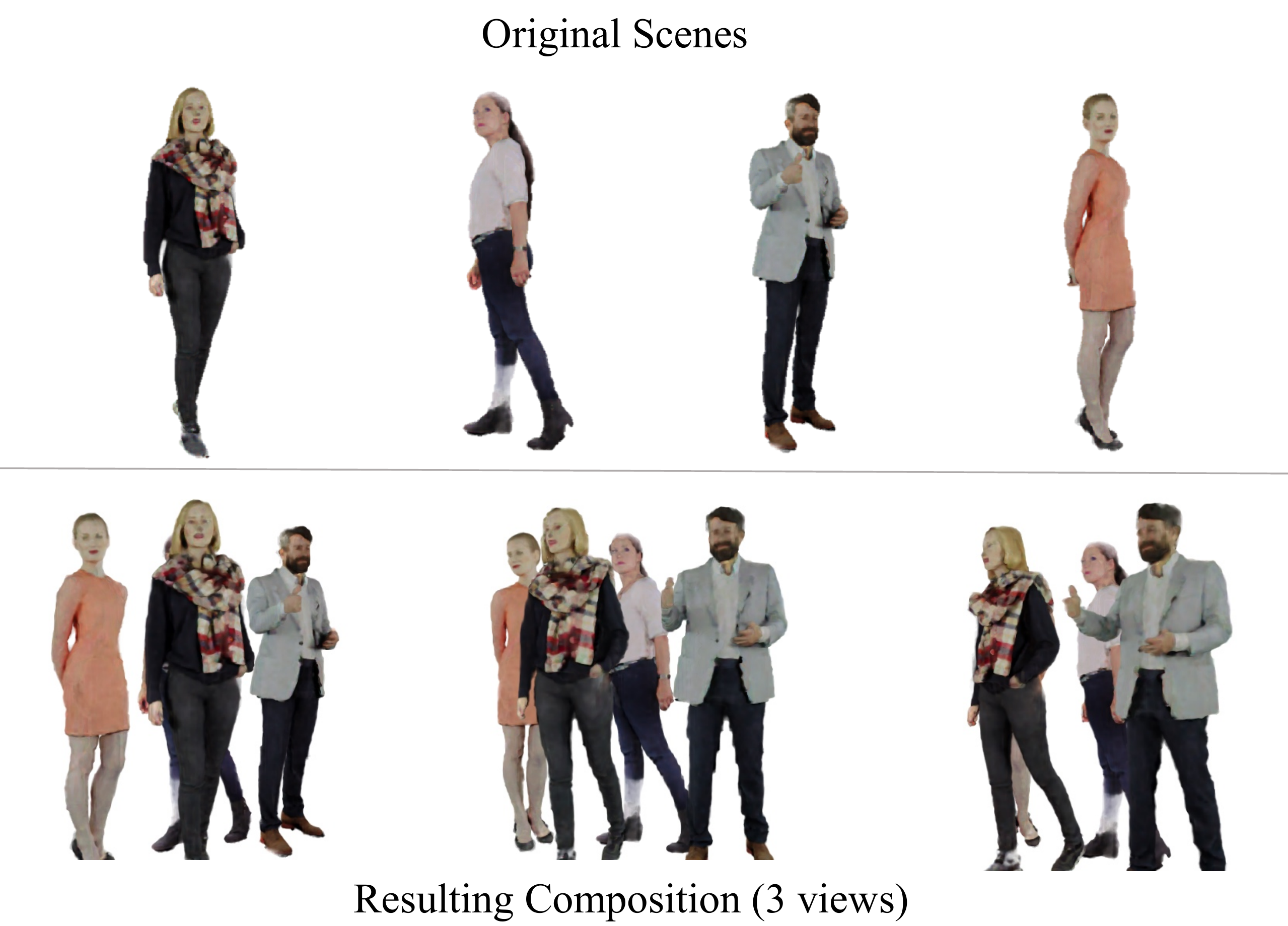}
  \captionof{figure}{\textbf{Combining people from AXYZ~\cite{axyz} dataset}. Note that we use the rendering network trained on 6 scenes from LLFF~\cite{mildenhall2019llff} and finetune a volume for each subject separately.
  }
  \label{fig:crowd}
\end{figure}

\begin{figure}
 \vspace{-1.0em}
  \centering
  \includegraphics[width=\columnwidth]{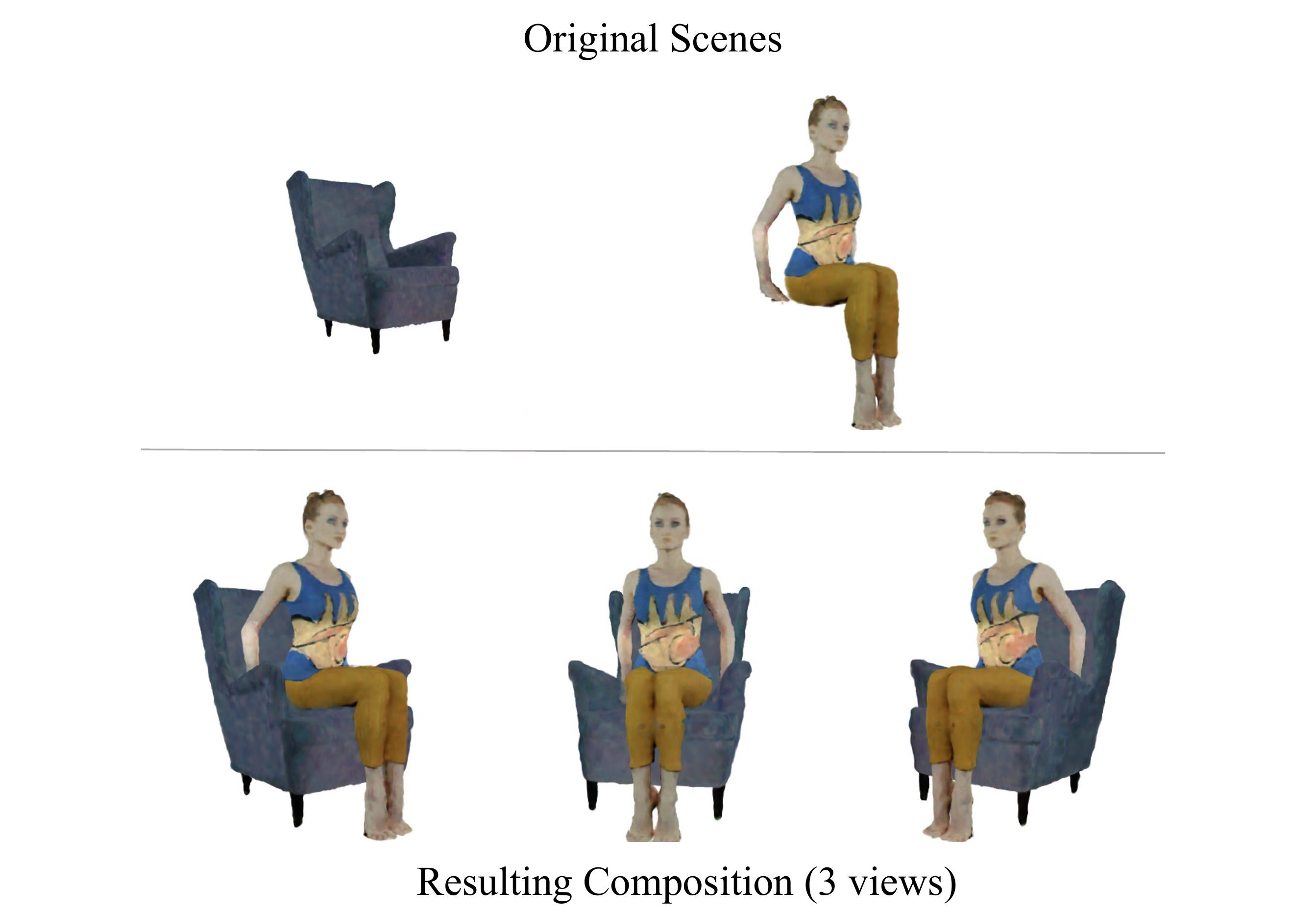}
  \captionof{figure}{\textbf{Combining objects from various datasets}. Tight object fusion is achieved by choosing in each voxel the feature set with maximum $L_2$ norm among voxels of original scenes.
  }
  \vspace{-1.5em}
  \label{fig:woman_chair}
\end{figure}

To demonstrate the generalization capacity of our networks, we provide additional results showing interactive scene composition, editing and novel view synthesis on multiple datasets from different domains.
For these results, we use the same network used for our previous experiments, trained on the $6$ LLFF scenes previously described.
The feature volumes for each subject are then optimized using the same approach as before, for the LLFF scenes withheld during training.

We use images from the DeepVoxels~\cite{sitzmann2019deepvoxels} dataset, which contains multiple calibrated images of static 3D objects such as furniture. We use $479$ images captured from the full $360^{\circ}$ field around an object, at a resolution of $512 \times 512$.
We also use a multi-view dataset set of similarly rendered images of textured 3D models of scanned human subjects from AXYZ Design~\cite{axyz} (25 primarly frontal images per subject, at a resolution of $512 \times 512$).
The feature volume optimization takes approximately $6$ hours per subject.

Fig.~\ref{fig:composition} portrays the combination of the feature volumes for a vase and a stand with that of a human subject from the AXYZ dataset.
We also we combine feature volumes for $4$ human subjects (Fig.~\ref{fig:crowd}).
Interestingly, despite the large difference in the appearance of the subjects in these datasets from the network training images, including a complete lack of humans in the LLFF images, and the relatively small number of scenes used for training the rendering network, the results are quite reasonable.
This suggests that the initial network parameter training and feature volume optimization does indeed learn a disentangled representation that allows for a flexible approach to rendering novel content beyond that which is similar to what it has seen during training.

We also noticed that the color and density information carried by each voxel correlates with the $L_2$ norm of its feature vector. This allows us to fuse feature volumes by choosing feature vector with maximum $L_2$ norm among the voxels with the same coordinates from original scenes. This way, we can achieve tight contact between objects from different scenes without artifacts (Fig.~\ref{fig:woman_chair}).

\begin{figure}[ht]
 \vspace{-1.0em}
  \begin{center}
    \includegraphics[width=0.44\textwidth]{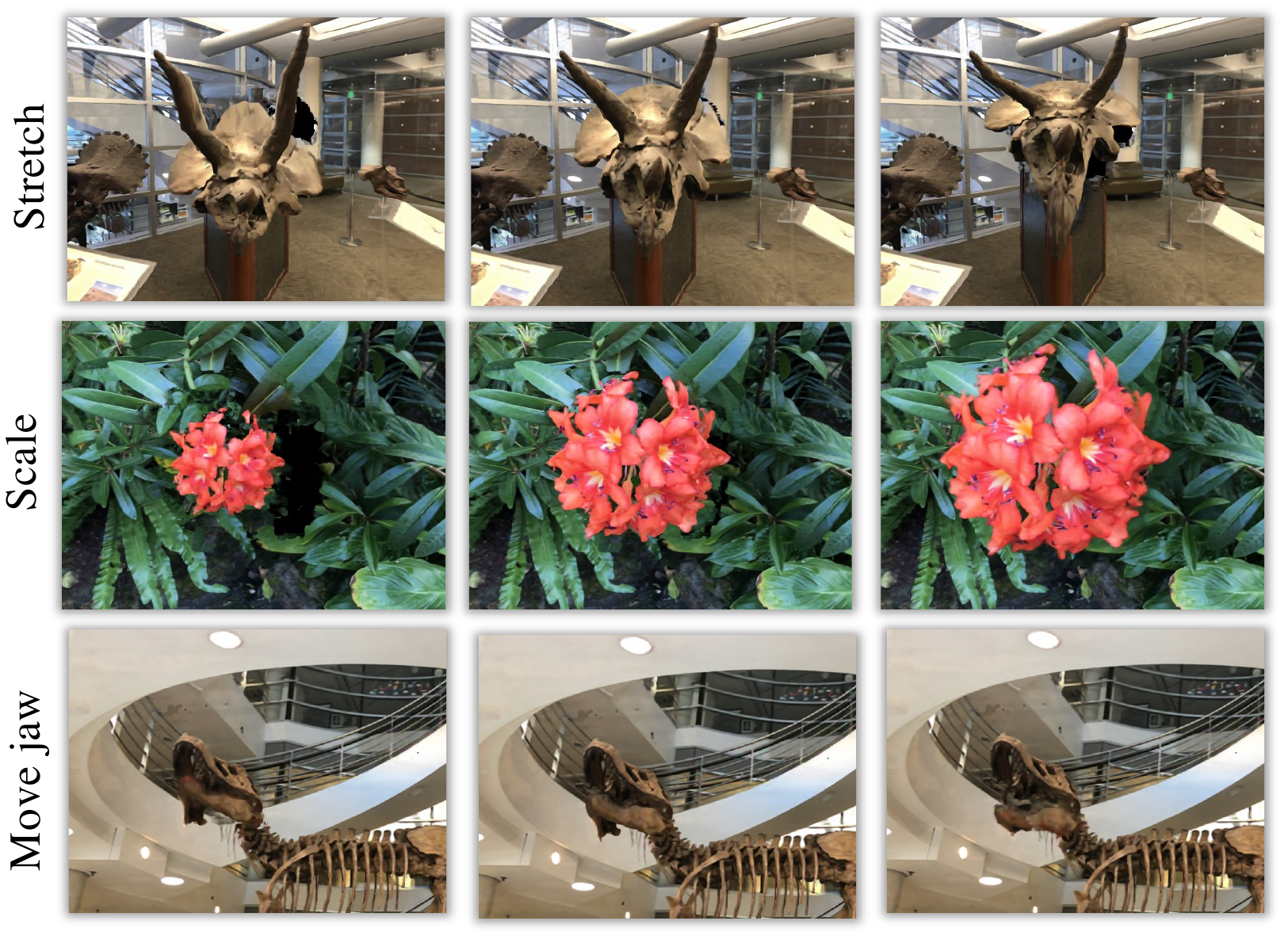}
  \end{center}
  \vspace{-1em}
  \captionof{figure}{\textbf{Rigid and non-rigid transformations} of objects extracted from a scene. The middle column shows the original object/scene. Please zoom in and consult the supplementary video for further demonstrations.} 
  \vspace{-1.0em}
  \label{fig:rigid}
\end{figure}

\vspace{-0.5em}
\subsection{User Study}
\vspace{-0.5em}
We also conducted a user study to evaluate the scene manipulation capabilities of our method and Neural Point-Based Graphics (NPBG)~\cite{aliev2019neural}.
Given $6$ scenes and $10$ pairs of edited images per scene (ours vs. NPBG), Amazon Mechanical Turk users were presented these pairs and asked to decide which image was preferable. $5$ workers were asked per image pair, for a total of $300$ questions asked ($18$ unique users participated). In $62\%$ of the cases, users preferred our edited images.

\vspace{-0.5em}
\subsection{Scene Content Deformation}
\vspace{-0.5em}
Fig.~\ref{fig:rigid} shows various rigid and non-rigid manipulations of objects extracted from these volumes and on entire scenes, obtained using the aforementioned volume deformation and resampling techniques.
While we show examples such as stretching and scaling specific scene content, the flexiblity of our editing framework allows for arbitrary manipulations that can be specified as local or global modifications to a scene's feature volume.

\vspace{-0.5em}
\subsection{Additional Video Results}
\vspace{-0.5em}
In the supplementary video, we show continuous novel view synthesis results, for both scenes used to train the rendering network, and for novel scenes for which we simply optimize the feature volumes.
We also show results for manipulated scenes in which content is combined from multiple training and novel scenes, as well as videos showing various creative manipulations such as the aforementioned local scaling and non-rigid deformation of the image content.

\begin{table}
\vspace{-1.5em}
\begin{center}
    \footnotesize
	\tabcolsep=0.11cm
	\begin{tabular}{|l||c|c|c|} \hline 
	
	 & PSNR$\uparrow$ & SSIM$\uparrow$ & LPIPS$\downarrow$ \\ \hline
	 NeRF~\cite{mildenhall2020nerf}&\textbf{28.045}&\textbf{0.881}&\textbf{0.137} \\ \hline 
	 NeRF-multiscene & 25.262 & 0.815 & 0.194 \\ \hline 
     Ours & 25.635 & 0.853 & 0.181 \\ \hline 
	  
	\end{tabular}
\end{center}
\vspace{-1em}
\caption{\textbf{Quantitative comparison with NeRF~\cite{mildenhall2020nerf}}. Metrics are averaged across test images for 8 scenes from from LLFF~\cite{mildenhall2019llff} dataset. Our method is trained on multiple scenes simultaneously while NeRF memorizes only one scene.}
\label{tab:nerf_comparison}
\vspace{-1.5em}
\end{table}

\begin{table*}
\vspace{-1.5em}
\begin{center}
    \footnotesize
    \tabcolsep=0.11cm
    \resizebox{\textwidth}{!}{
    \begin{tabular}{l|c|c|c|c|c|c||c||c|c||c||c} %
      \multicolumn{12}{c}{PSNR$\uparrow$} \\
      \multirow{2}{*}{}& \multicolumn{7}{|c||}{Per-Scene, Training} & \multicolumn{3}{|c||}{Per-Scene, Novel} & \multicolumn{1}{|c}{}\\
      & Room & Leaves & Fortress & Orchids & Flower & Horns & Avg. & Fern & T-Rex & Avg. & Total Avg. \\ \hline
      NeRF~\cite{mildenhall2020nerf} & \textbf{33.965} & \textbf{22.562} & \textbf{33.099} & \textbf{21.276} & \textbf{28.564} & \textbf{29.484} & \textbf{28.158} & \textbf{26.843} & \textbf{28.567} & \textbf{27.705} & \textbf{28.045} \\ \hline
      Ours & 30.938 & 18.438 & 28.930 & 21.182 & 27.526 & 25.807 & 25.470 & 25.752 & 26.510 & 26.131 & 25.635 \\ \hline
    \end{tabular}
    }
\end{center}
\begin{center}
    \footnotesize
    \tabcolsep=0.11cm
    \resizebox{\textwidth}{!}{
    \begin{tabular}{l|c|c|c|c|c|c||c||c|c||c||c} %
      \multicolumn{12}{c}{SSIM$\uparrow$} \\
      \multirow{2}{*}{}& \multicolumn{7}{|c||}{Per-Scene, Training} & \multicolumn{3}{|c||}{Per-Scene, Novel} & \multicolumn{1}{|c}{}\\
      & Room & Leaves & Fortress & Orchids & Flower & Horns & Avg. & Fern & T-Rex & Avg. & Total Avg. \\ \hline
      NeRF~\cite{mildenhall2020nerf} & \textbf{0.963} & \textbf{0.814} & \textbf{0.932} & 0.744 & \textbf{0.893} & \textbf{0.912} & \textbf{0.876} & \textbf{0.856} & \textbf{0.930} & \textbf{0.893} & \textbf{0.881} \\ \hline
      Ours & 0.943 & 0.770 & 0.861 & \textbf{0.764} & 0.883 & 0.875 & 0.849 & 0.820 & 0.907 & 0.864 & 0.853 \\ \hline
    \end{tabular}
    }
\end{center}
\begin{center}
    \footnotesize
    \tabcolsep=0.11cm
    \resizebox{\textwidth}{!}{
    \begin{tabular}{l|c|c|c|c|c|c||c||c|c||c||c} %
      \multicolumn{12}{c}{LPIPS$\downarrow$} \\
      \multirow{2}{*}{}& \multicolumn{7}{|c||}{Per-Scene, Training} & \multicolumn{3}{|c||}{Per-Scene, Novel} & \multicolumn{1}{|c}{}\\
      & Room & Leaves & Fortress & Orchids & Flower & Horns & Avg. & Fern & T-Rex & Avg. & Total Avg. \\ \hline
      NeRF~\cite{mildenhall2020nerf} & \textbf{0.093} & \textbf{0.186} & \textbf{0.068} & 0.204 & \textbf{0.110} & \textbf{0.133} & \textbf{0.132} & \textbf{0.168} & \textbf{0.137} & \textbf{0.152} & \textbf{0.137} \\ \hline
      Ours & 0.131 & 0.227 & 0.207 & \textbf{0.178} & 0.123 & 0.190 & 0.176 & 0.236 & 0.153 & 0.195 & 0.181 \\ \hline
    \end{tabular}
    }
\end{center}
\caption{\textbf{Per-scene quantitative results compared with the results of the original NeRF implementation}. We report the results with networks trained per-scene until convergence using their approach, while ours, uses 1 network trained for all scenes.}
\label{tab:tab_nerf_compare}
\end{table*}

\subsection{Comparisons with NeRF}

\paragraph{Multi-scene NeRF.}
In Table~\ref{tab:nerf_comparison} we show a comparison to NeRF \cite{mildenhall2020nerf} on NVS. As NeRF is a scene-specific method, while our rendering network generalizes across scenes, we adapted it for a multi-scene scenario by associating each scene with a one-hot encoding vector.
NeRF is then conditioned on this code in order to generate the specific scene. We have increased the capacity of the network accordingly to accommodate multiple scenes. While the original NeRF performs better than our method, we outperform NeRF in the multiscene scenario.

\vspace{-0.5em}
\paragraph{Single-scene NeRF.}
Below we provide further details on the comparisons with the original implementation of NeRF~\cite{mildenhall2020nerf}, using the training parameters described in Sec.~\ref{sec:supp_training} of this document on the LLFF dataset.
For these comparisons, we measure the difference between synthesized novel views and ground-truth images withheld for each scene during training, as in their evaluations.
In our experiments, the total amount of computation time required to optimize the radiance function parameters for NeRF for a single scene until convergence, which required approximately $48$ hours, or roughly $2$ days.~\footnote{These numbers are slightly different than those reported in~\cite{mildenhall2020nerf}, but as noted above, we use a different training image resolution and number of samples in the fine network $\Cfine$ in our experiments. As pre-trained models for each of these scenes were not available for their implementation, we trained the models for each scene using the above parameters for a more direct comparison.}
Thus, training for all of the $8$ scenes in the results depicted in Tab.~\ref{tab:tab_nerf_compare}, a total of approximately $16$ days of computation was required (though this was performed in parallel on multiple systems) using an NVIDIA V100 GPU for each scene.

In contrast, training our \emph{single} rendering network on the set of $6$ training scenes took a total of roughly $36$ hours. This network can then be applied to novel scenes using our feature volume optimization process.
For the novel scenes, an average of $5.5$ hours (also performed in parallel on multiple systems) was required to compute their feature volumes, meaning that for these $8$ scenes in total roughly $47$ hours, or slightly less than $2$ days of total computation was required.
We further note that, while NeRF essentially memorizes a representation of the training scene that allows for a limited range of novel view synthesis, our approach additionally allows for the intuitive manipulation and combinination of data from multiple scenes, as demonstrated in our experimental results.
Thus, given that computational efficiency in training a single network for multiple scenes networks is an advantage of our approach, we conduct additional %
evaluations in which we examine how well NeRF performs with similar computational resources.

We trained the NeRF network for each of the $8$ scenes for 100K iterations, or approximately $5.5$ hours, which is comparable to the time required to run the novel scene feature volume optimization for a single scene using our approach. After this point, in our experiments the performance improved slowly until converging after approximately $2$ days to the aforementioned results.
The results are depicted in Tab.~\ref{tab:tab_compare}, with the results for Neural Point-Based Graphics (NPBG)~\cite{aliev2019neural} included for further comparison.
As seen in these tables, while NeRF does produce results that are slightly more visually appealing when given unrestricted computational resources, when the training time is restricted the results are comparable, with ours outperforming each alternative on average in $2$ out of $3$ metrics.
We also note that these $2$ metrics, SSIM and LPIPS, are generally regarded to correspond better to realistic and more higher quality images for the human visual system~\cite{5596999}.
Additionally, in Table~\ref{tab:nsvf_compare}, we show more detailed comparisons against Neural Point-Based Graphics (NPBG)~\cite{aliev2019neural} and Neural Sparse Voxel Fields~\cite{Liu20neurips_sparse_nerf}.

\begin{table*}
\begin{center}
    \footnotesize
    \tabcolsep=0.11cm
    \resizebox{\textwidth}{!}{
    \begin{tabular}{l|c|c|c|c|c|c||c||c|c||c||c} %
      \multicolumn{12}{c}{PSNR$\uparrow$} \\
      \multirow{2}{*}{}& \multicolumn{7}{|c||}{Per-Scene, Training} & \multicolumn{3}{|c||}{Per-Scene, Novel} & \multicolumn{1}{|c}{}\\
      & Room & Leaves & Fortress & Orchids & Flower & Horns & Avg. & Fern & T-Rex & Avg. & Total Avg. \\ \hline
      NPBG~\cite{aliev2019neural} & 26.058 & 17.854 & 19.172 & 17.535 & 22.106 & 20.651 & 20.563 & 18.285 & 20.407 & 19.346 & 20.259 \\ \hline
      NeRF~\cite{mildenhall2020nerf} 100K & \textbf{32.492} & \textbf{21.952} & \textbf{31.957} & 21.172 & 27.478 & \textbf{27.922} & \textbf{27.162} & \textbf{26.370} & \textbf{27.209} & \textbf{26.790} & \textbf{27.069} \\ \hline
      Ours & 30.938 & 18.438 & 28.930 & \textbf{21.182} & \textbf{27.526} & 25.807 & 25.470 & 25.752 & 26.510 & 26.131 & 25.635 \\ \hline
    \end{tabular}
    }
\end{center}
\begin{center}
    \footnotesize
    \tabcolsep=0.11cm
    \resizebox{\textwidth}{!}{
    \begin{tabular}{l|c|c|c|c|c|c||c||c|c||c||c} %
      \multicolumn{12}{c}{SSIM$\uparrow$} \\
      \multirow{2}{*}{}& \multicolumn{7}{|c||}{Per-Scene, Training} & \multicolumn{3}{|c||}{Per-Scene, Novel} & \multicolumn{1}{|c}{}\\
      & Room & Leaves & Fortress & Orchids & Flower & Horns & Avg. & Fern & T-Rex & Avg. & Total Avg. \\ \hline
      NPBG~\cite{aliev2019neural} & 0.890 & 0.668 & 0.804 & 0.572 & 0.769 & 0.794 & 0.750 & 0.706 & 0.774 & 0.740 & 0.747 \\ \hline
      NeRF~\cite{mildenhall2020nerf} 100K & \textbf{0.953} & \textbf{0.776} & \textbf{0.908} & 0.718 & 0.859 & 0.870 & 0.847 & \textbf{0.830} & 0.903 & \textbf{0.867} & 0.852 \\ \hline
      Ours & 0.943 & 0.770 & 0.861 & \textbf{0.764} & \textbf{0.883} & \textbf{0.875} & \textbf{0.849} & 0.820 & \textbf{0.907} & 0.864 & \textbf{0.853} \\ \hline
    \end{tabular}
    }
\end{center}
\begin{center}
    \footnotesize
    \tabcolsep=0.11cm
    \resizebox{\textwidth}{!}{
    \begin{tabular}{l|c|c|c|c|c|c||c||c|c||c||c} %
      \multicolumn{12}{c}{LPIPS$\downarrow$} \\
      \multirow{2}{*}{}& \multicolumn{7}{|c||}{Per-Scene, Training} & \multicolumn{3}{|c||}{Per-Scene, Novel} & \multicolumn{1}{|c}{}\\
      & Room & Leaves & Fortress & Orchids & Flower & Horns & Avg. & Fern & T-Rex & Avg. & Total Avg. \\ \hline
      NPBG~\cite{aliev2019neural} & 0.163 & 0.272 & 0.200 & 0.301 & 0.204 & 0.222 & 0.227 & 0.267 & 0.231 & 0.249 & 0.232 \\ \hline
      NeRF~\cite{mildenhall2020nerf} 100K & \textbf{0.126} & 0.235 & \textbf{0.107} & 0.250 & 0.155 & 0.199 & 0.179 & \textbf{0.212} & 0.178 & \textbf{0.195} & 0.183 \\ \hline
      Ours & 0.131 & \textbf{0.227} & 0.207 & \textbf{0.178} & \textbf{0.123} & \textbf{0.190} & \textbf{0.176} & 0.236 & \textbf{0.153} & \textbf{0.195} & \textbf{0.181} \\ \hline
    \end{tabular}
    }
\end{center}
\caption{\textbf{Full per-scene quantitative results}. We report the per-scene and average results on the initial training scenes as well as on the novel scenes used in our generalization process, as well as the average across both datasets.}
\label{tab:tab_compare}
\end{table*}

\begin{table*}
\centering
\begin{tabular*}{\textwidth}{c @{\extracolsep{\fill}} ccccccccc}
\multicolumn{9}{c}{PSNR $\uparrow$} \\
& Fern & Leaves & Fortress & Orchids & Flower & Trex & Horns & Average \\ \hline
NSVF & 20.594 & 17.316 & 26.901 & 14.309 & 22.930 & 17.467 & 23.380 & 20.414 \\ \hline
NPBG & 18.285 & 17.854 & 19.172 & 17.535 & 22.106 & 20.407 & 20.651 & 19.430  \\ \hline
Ours & \textbf{25.752} & 18.438 & 28.930 & \textbf{21.182} & \textbf{27.526} & \textbf{26.510} & \textbf{25.807} & \textbf{24.878}  \\ \hline
Ours (single scene) & 25.082 & \textbf{20.554} & \textbf{29.618} & 20.374 & 26.260 & 24.753 & 25.425 & 24.581 \\ \hline \\
\end{tabular*}

\begin{tabular*}{\textwidth}{c @{\extracolsep{\fill}} ccccccccc}
\multicolumn{9}{c}{SSIM $\uparrow$} \\
& Fern & Leaves & Fortress & Orchids & Flower & Trex & Horns & Average \\ \hline
NSVF & 0.575 & 0.402 & 0.721 & 0.250 & 0.629 & 0.490 & 0.682 & 0.536 \\ \hline
NPBG & 0.706 & 0.668 & 0.804 & 0.572 & 0.769 & 0.774 & 0.794 & 0.727 \\ \hline
Ours & \textbf{0.820} & \textbf{0.770} & \textbf{0.861} & \textbf{0.764} & \textbf{0.883} & \textbf{0.907} & \textbf{0.875} & \textbf{0.840} \\ \hline
Ours (single scene) & 0.792 & 0.738 & 0.854 & 0.704 & 0.840 & 0.868 & 0.834 & 0.804 \\ \hline
\end{tabular*}

\begin{tabular*}{\textwidth}{c @{\extracolsep{\fill}} ccccccccc}
\multicolumn{9}{c}{LPIPS $\downarrow$} \\
& Fern & Leaves & Fortress & Orchids & Flower & Trex & Horns & Average \\ \hline
NSVF & 0.448 & 0.519 & 0.346 & 0.571 & 0.385 & 0.445 & 0.431 & 0.449 \\ \hline
NPBG & 0.267 & 0.272 & 0.200 & 0.301 & 0.204 & 0.231 & 0.222 & 0.242 \\ \hline
Ours & \textbf{0.236} & \textbf{0.227} & 0.207 & \textbf{0.178} & \textbf{0.123} & \textbf{0.153} & \textbf{0.190} & \textbf{0.188} \\ \hline
Ours (single scene) & 0.272 & 0.261 & \textbf{0.198} & 0.245 & 0.173 & 0.195 & 0.238 & 0.226 \\ \hline \\
\end{tabular*}
\vspace{0.5em}
\caption{\textbf{Quantitative comparison with NPBG~\cite{aliev2019neural} and NSVF~\cite{Liu20neurips_sparse_nerf}}. Metrics are computed across test images for scenes from from LLFF~\cite{mildenhall2019llff} dataset. ``Ours'' is our method trained on 6 scenes simultaneously as in our original setup. ``Ours (single scene)'' is our method trained for one scene at a time.}
\label{tab:nsvf_compare}
\end{table*}